\documentclass{article}

\usepackage[main, final]{neurips_2026}

\usepackage[utf8]{inputenc} 
\usepackage[T1]{fontenc}    
\usepackage{hyperref}       
\usepackage{url}            
\usepackage{booktabs}       
\usepackage{amsfonts}       
\usepackage{nicefrac}       
\usepackage{microtype}      
\usepackage{xcolor}         
\usepackage{multirow}
\usepackage{graphicx}
\usepackage{wrapfig}
\usepackage{booktabs}
\usepackage{multirow}
\usepackage{graphicx}
\usepackage{makecell}
\usepackage{subcaption}
\usepackage{amsmath}

\usepackage{tcolorbox}
\usepackage[table]{xcolor}
\definecolor{tabcolor3}{rgb}{0.86, 0.93, 1.0}
\newcommand{\method}{SSM }

\title{Accelerating Diffusion Language Models via Structured Suffix Modeling}

\author{%
  Zifeng Cheng$^\text{\textnormal{1}}$\thanks{Equal contribution.} \quad 
  Keda Li$^\text{\textnormal{1}}$\footnotemark[1]  \quad 
 Zhiwei Jiang$^\text{\textnormal{1}}$\thanks{Corresponding author.}  \quad 
  \textbf{Cong Wang}$^\text{\textnormal{1}}$ \quad
 \textbf{Fei Shen}$^\text{\textnormal{2}}$  \quad
 \textbf{Qing Gu}$^\text{\textnormal{1}}$ \\
 $^\text{1}$ State Key Laboratory for Novel Software Technology, Nanjing University \\
$^\text{2}$ National University of Singapore
\\
  \texttt{chengzf@nju.edu.cn}, \texttt{522026320069@smail.nju.edu.cn}, \texttt{jzw@nju.edu.cn}, \\
  \texttt{wang.c@nju.edu.cn}, \texttt{shenfei29@nus.edu.sg},  \texttt{guq@nju.edu.cn}
}

\begin{document}

\maketitle

\begin{abstract}
Diffusion Language Models (DLMs) exhibit strong parallel decoding capabilities by denoising multiple tokens in a single generation step.
However, this parallelism comes with substantial computational overhead, as each step requires interactions with all suffix tokens.
Existing methods typically reduce this cost by retaining only a local suffix window as a substitute for the full suffix.
Despite their effectiveness, these methods overlook the structural heterogeneity across suffix regions and re-initialize suffix tokens with identical representations at each timestep.
To this end, we propose a structured suffix modeling method for efficient DLM inference.
Specifically, we divide the suffix into three regions, i.e., the local, middle, and tail regions, and retain different numbers of suffix tokens in each region according to their structural roles.
Moreover, we incorporate the decoding results from the previous step into the suffix token representations at the current step, allowing them to carry evolving denoising information across generation steps.
Notably, our method is training-free and orthogonal to several existing acceleration techniques, such as parallel decoding strategies and KV cache.
Empirical results across multiple benchmarks on three DLMs demonstrate that our method can further accelerate DLM inference and improve performance in most cases.
In particular, in long-sequence inference, our method achieves up to a \(72.81\times\) speedup when combined with other acceleration techniques.
Our code is available at \url{https://github.com/zifengcheng/SSM}.
\end{abstract}

\section{Introduction}
Diffusion Language Models (DLMs)~\citep{DBLP:conf/nips/LiTGLH22,DBLP:conf/acl/HeSTWHQ23,DBLP:conf/nips/ShiHWDT24,llada,dream} offer a new paradigm for parallel decoding over entire sequences or blocks by leveraging bidirectional context.
Compared to autoregressive decoding~\citep{qwen2.5,dubey2024llama}, this paradigm can predict multiple tokens within a single generation step and effectively reduce latency.
However, in block-wise decoding~\citep{llada,dream}, each step not only denoises the current block but also computes attention scores and hidden states for a large number of masked suffix tokens.
Although these suffix tokens provide future contextual signals, their associated computation accounts for a substantial portion of the overall cost, limiting the decoding speed.
As the generation length increases, this overhead becomes increasingly pronounced, making suffix computation a critical bottleneck for efficient DLM inference.


Recently, several studies~\citep{chen2026dpad,Streaming-dLLM} have explored accelerating DLM inference by directly excluding redundant suffix tokens from the model input.
Unlike conventional attention-score pruning methods~\citep{wang21,DBLP:conf/aaai/SongLLLHGHQ26}, which require computing attention scores before pruning tokens based on their magnitudes, suffix dropout predefines the sparsity pattern before the input is fed into the model, without accessing internal model states.
DPad~\citep{chen2026dpad} presents the first systematic investigation into suffix tokens, revealing that they primarily function as a non-semantic information reservoir.
Motivated by this observation, DPad retains tokens only within a local suffix window and applies distance-aware dropout to determine whether each token in the window should be preserved.
Streaming-dLLM~\citep{Streaming-dLLM} preserves neighboring suffix tokens and the final token to maintain essential contextual and boundary cues, while pruning redundant suffix tokens.

Despite their effectiveness, existing methods still largely formulate suffix dropout as a locality-oriented pruning problem. 
First, these methods fail to explicitly model the functional roles of different suffix blocks during block-wise denoising. 
Tokens located in different suffix blocks may provide distinct contextual, structural, and boundary cues, and thus should be assigned different token budgets. 
Second, suffix tokens are re-initialized at each timestep without incorporating information from previous timesteps, thereby discarding useful denoising signals accumulated during iterative refinement. 
As a result, existing methods may fail to preserve informative suffix cues and exploit the evolving information propagated across generation steps.


To address these limitations, we propose SSM, a structured suffix modeling framework for efficient DLM inference.
Our key insight is that suffix tokens in different regions serve different roles during block-wise decoding.
Specifically, the attention scores received by suffix tokens first drop rapidly as the token index increases, then remain stable, and finally rise again.
Thus, we first divide the suffix into three regions: the local region, the middle region, and the tail region.
For the local region, we retain all neighboring suffix tokens to preserve essential contextual information.
For the middle region, we keep only the start token of each suffix block as a lightweight structural anchor, maintaining the coarse suffix skeleton with minimal computation.
For the tail region, we additionally retain the final token to explicitly preserve terminal boundary information.
Furthermore, we introduce a soft suffix embedding mechanism.
At each timestep, we incorporate the decoding results from the previous step into the suffix token representations at the current step.
In this way, suffix tokens can carry evolving denoising information across generation steps.

Our method is training-free and can be directly applied to existing DLMs at inference time.
Moreover, it is orthogonal to existing acceleration techniques such as parallel decoding and KV cache, enabling further efficiency gains when combined with them.
We conduct extensive experiments on mathematical reasoning and code generation benchmarks using LLaDA-Instruct, LLaDA-1.5, and Dream-Base.
Empirical results show that \method consistently reduces inference latency while maintaining competitive accuracy.
Notably, in long-sequence inference, our method achieves up to a \(72.81\times\) speedup when combined with other acceleration techniques.

Our contributions are summarized as follows:
\begin{itemize}
    \item We reformulate suffix dropout in DLMs as a role-aware structured suffix modeling problem, revealing that suffix tokens in different regions provide distinct contextual, structural, and boundary cues rather than serving as a homogeneous redundancy reservoir.
    \item We introduce a soft suffix embedding mechanism that incorporates previous-step decoding results into current suffix token representations, enabling suffix tokens to carry evolving denoising information across generation steps.
    \item We develop a training-free inference acceleration method that is orthogonal to existing acceleration techniques, achieving efficient DLM inference with competitive or improved task performance.
\end{itemize}

\section{Related work}

\paragraph{Accelerating Diffusion Language Models}
Unlike autoregressive models that generate one token at a time, dLLMs leverage bidirectional attention to denoise multiple tokens within a single timestep, thereby enabling parallel decoding.
Recently, many studies have been devoted to accelerating the inference process of dLLMs to fully exploit their parallel decoding capabilities.
Among them, decoding strategies~\citep{wu2026fastdllm,kim2025klass,huang2025pc,ben-hamu2025accelerated,kim2025train} and KV cache~\citep{ma25dkv,liu2025dllm,hu2026flashdlm,DBLP:conf/aaai/SongLLLHGHQ26} are two of the most representative categories of methods.
Decoding methods focus on defining new decoding criteria to accelerate generation, with Fast-dLLM~\citep{wu2026fastdllm} as a representative approach that performs decoding based on confidence thresholds.
In addition, some studies~\citep{latent,zhong2026beyond} have explored latent refinement decoding. Unlike these methods, our approach focuses on scenarios with only a small number of suffix tokens and does not require iterative refinement.
KV cache methods accelerate inference by caching the representations of a subset of tokens to avoid redundant computation.
Notably, these methods are orthogonal to our approach and can be combined with it to further improve inference efficiency.


\paragraph{Suffix Dropout}
Recently, some studies~\citep{chen2026dpad,Streaming-dLLM} have explored accelerating DLM inference by directly excluding a subset of suffix tokens from the model input.
Unlike conventional sparsification strategies \citep{DBLP:conf/aaai/SongLLLHGHQ26,wang2026sparsed} that require access to the internal states of DLMs, such as attention weights, suffix dropout can discard suffix tokens in advance without accessing internal model states.

DPad~\citep{chen2026dpad} systematically investigates suffix tokens and reveals that they primarily function as a non-semantic information reservoir that aggregates signals propagated from prefix tokens across multiple Transformer layers.
Motivated by this observation, DPad defines a local window and applies distance-aware token dropout within the window at the beginning of decoding each block.
Streaming-dLLM~\citep{Streaming-dLLM} preserves neighboring suffix tokens and the final token to maintain essential contextual and boundary cues while pruning redundant suffix tokens, and further introduces an orthogonal dynamic confidence-aware parallel decoding strategy.
This parallel decoding mechanism adaptively adjusts decoding thresholds based on both the diffusion stage and the intra-block confidence distribution.
However, these methods overlook the heterogeneity across suffix regions and keep suffix token representations unchanged throughout inference.

\section{Method}


\begin{figure}[t]
    \centering
    \begin{subfigure}[b]{0.45\textwidth}
        \includegraphics[width=\textwidth]{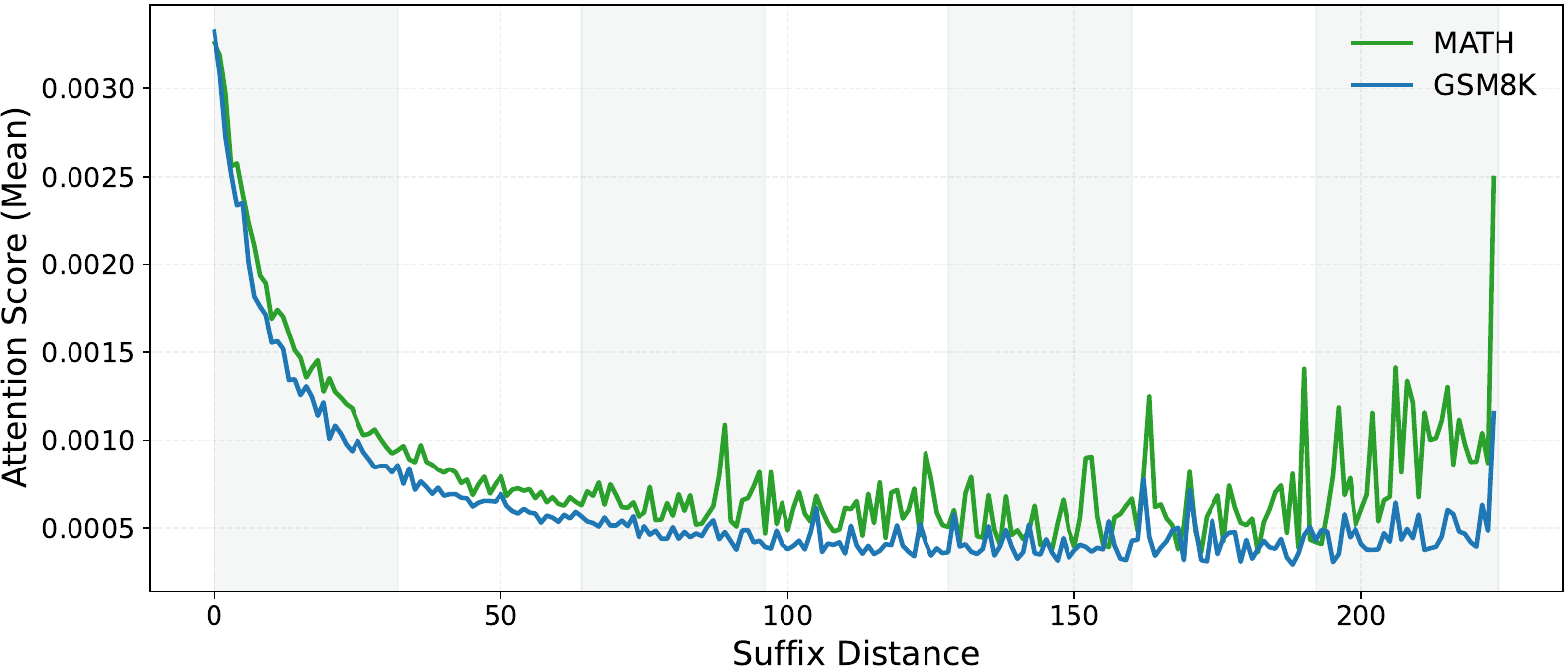}
    \end{subfigure}
    \begin{subfigure}[b]{0.45\textwidth}
        \includegraphics[width=\textwidth]{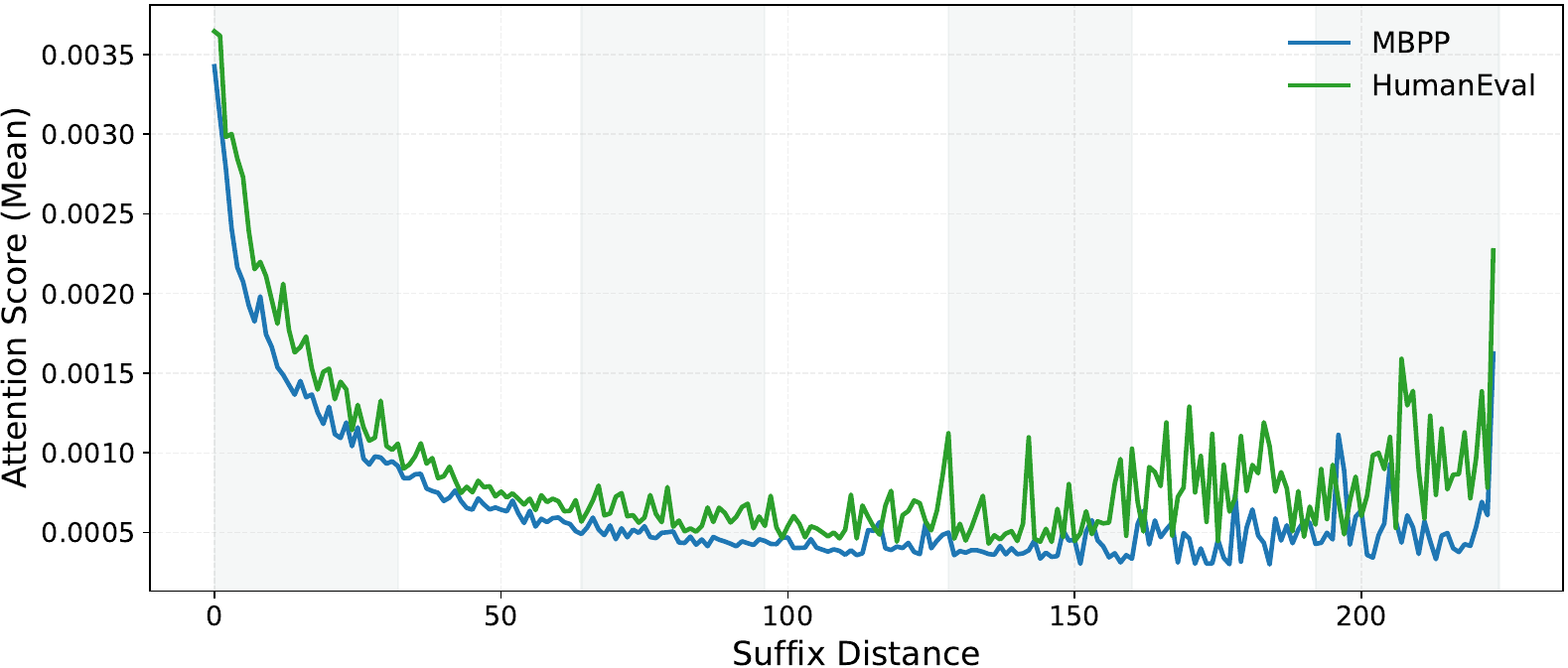}
    \end{subfigure}
    \caption{Average attention scores assigned to suffix tokens across all attention heads on four datasets using LLaDA-Instruct, with the generation length set to 256 and the block size set to 32.
}
    \label{fig:motivation}
\end{figure}

\subsection{Motivation}
To investigate whether all suffix tokens contribute equally to block-wise denoising, we visualize the average attention scores received by suffix tokens across all attention heads in Figure~\ref{fig:motivation}.
The results reveal a clear region-dependent pattern: attention scores initially decrease, then plateau, and finally rise again near the tail region.
Nearby suffix tokens receive substantially higher attention scores, suggesting that they provide fine-grained local context for the current decoding block.
As the suffix distance increases, the attention scores drop rapidly, indicating that dense interactions with distant suffix tokens are largely redundant.
Nevertheless, the attention scores increase sharply in the final block, suggesting that terminal boundary information remains important for decoding the current block.

These observations indicate that suffix tokens should not be handled by a uniform dropout rule.
Instead, different suffix regions should be assigned different token budgets according to their functional roles.
Specifically, regions close to the current decoding block are the most important and should retain more tokens to preserve local contextual information.
The final block is also relatively important for maintaining terminal boundary cues.
In contrast, middle suffix blocks are relatively less important and can be represented with a much smaller token budget.

\subsection{Structured Suffix Modeling}

\begin{figure}[t]
  \centering
  \includegraphics[width=\textwidth]{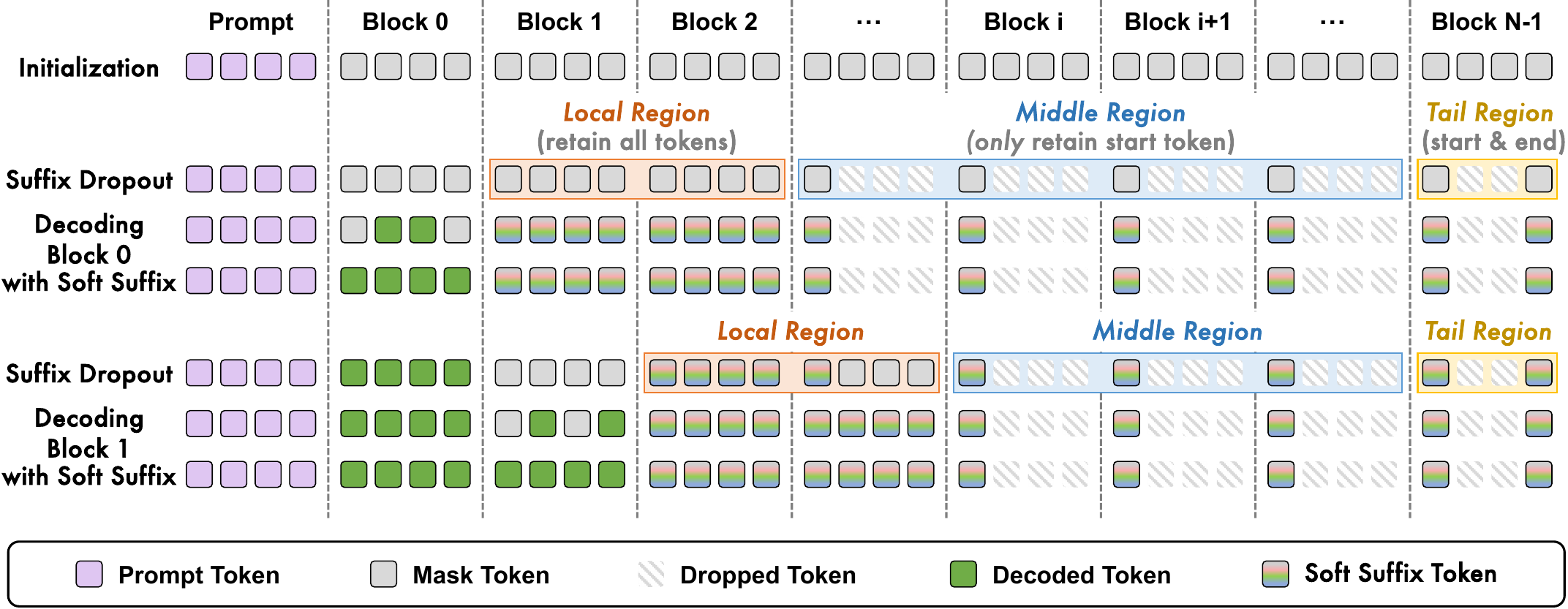}
  \caption{Overview of the proposed structured suffix modeling framework. The method divides the suffix into three regions and retains suffix tokens separately for each region.}
  \label{fig:overview}
\end{figure}

Motivated by the above observations, we propose a training-free Structured Suffix Modeling (SSM) framework for efficient DLM inference, as shown in Figure~\ref{fig:overview}.
The core idea of SSM is to model the suffix as a structured information field rather than a homogeneous set of redundant masked tokens. 

Specifically, SSM consists of two key components.
First, we decompose the suffix into three functional regions: a local region that carries fine-grained contextual information, a middle region that provides lightweight structural anchors, and a tail region that preserves terminal boundary information.
Different token budgets are then assigned to these regions according to their functional roles, reducing redundant suffix computation while preserving the key information required for block-wise denoising.
Second, we introduce a soft suffix embedding mixing mechanism, which incorporates previous-step decoding results into current suffix token representations.
This enables retained suffix tokens to carry evolving denoising signals across generation steps, instead of being repeatedly initialized as identical \texttt{[MASK]} embeddings.
Finally, the resulting sparsified input is fed into the DLM for decoding.

\paragraph{Suffix Region Partition}
Since suffix tokens in different regions have varying importance, we first divide the suffix into three regions, namely the local region, the middle region, and the tail region, and assign different retention ratios to different regions.

We first partition the suffix blocks \(S_t\) at timestep \(t\) into three regions:
\begin{equation}
S_t = S_t^{\mathrm{local}} \cup S_t^{\mathrm{middle}} \cup S_t^{\mathrm{tail}}.
\end{equation}
Specifically, the first \(w\) suffix blocks are defined as the local region, the last suffix block is defined as the tail region, and the remaining blocks between them are defined as the middle region, where \(w\) controls the size of the local region.
In particular, when the number of suffix blocks is less than \(w\), all suffix blocks are assigned to the local region.

\paragraph{Suffix Token Retention}
After suffix region partitioning, we apply different token retention strategies to the three regions.
For the local region, since it is close to the current decoding block and receives higher attention scores, we retain all tokens in the local region to preserve short-range contextual dependencies:
\begin{equation}
\mathcal{R}_t^{\mathrm{local}}
=
S_t^{\mathrm{local}}
\end{equation}
where $\mathcal{R}_t^{\mathrm{local}}$ denotes the tokens retained in the local region.

For the middle region, since it receives lower attention scores, we retain only the start token of each block to preserve the skeletal structure of the middle region.
\begin{equation}
\mathcal{R}_t^{\mathrm{middle}}
=
\{ \mathrm{start}(B_j) \mid B_j \in S_t^{\mathrm{middle}} \}.
\end{equation}
where $B_j$ denotes $j$-th block and start($\cdot$) denotes the start-token selection function.
Notably, we empirically find that the start token in each middle block usually receives higher attention scores than other suffix tokens, possibly because it is closer to the current decoding block.
Therefore, we choose to retain the start token as a simple and effective strategy.

For the tail region, in addition to the start token, we further retain the final token:
\begin{equation}
\mathcal{R}_t^{\mathrm{tail}} =
\{ \mathrm{start}(B_{\mathrm{tail}}), \mathrm{end}(B_{\mathrm{tail}}) \}.
\end{equation}
The final token provides important boundary information, allowing the DLM to plan the future generation length from the beginning of decoding.

Finally, we merge all tokens retained from the three regions to form the suffix set retained at step $t$:
\begin{equation}
\mathcal{R}_t =
\mathcal{R}_t^{\mathrm{local}}
\cup
\mathcal{R}_t^{\mathrm{middle}}
\cup
\mathcal{R}_t^{\mathrm{tail}}.
\end{equation}

\paragraph{Soft Suffix Embedding Mixing}
To enable suffix tokens to retain prediction information from previous timesteps, we introduce a soft suffix embedding mixing mechanism.
Specifically, we mix the embeddings of previously decoded tokens into the \texttt{[MASK]} representations, allowing suffix tokens to carry evolving denoising information across generation steps.

For each retained suffix position \(i \in \mathcal{R}_t\), let \(p_{t+1}^i(v)\) denote the predicted probability of token \(v\) at the previous timestep \(t+1\).
We select the top-\(k\) candidate tokens according to \(p_{t+1}^i\), denoted as \(\mathcal{T}_{t+1}^i\), and re-normalize their probabilities within this set:
\begin{equation}
\tilde{p}_{t+1}^i(v)
=
\frac{p_{t+1}^i(v)}
{\sum_{u \in \mathcal{T}_{t+1}^i} p_{t+1}^i(u)},
\quad v \in \mathcal{T}_{t+1}^i .
\end{equation}
Given the token embedding \(\mathbf{e}_v\) for token \(v\), we construct the soft decoded embedding as a weighted combination of the top-\(k\) token embeddings:
\begin{equation}
\tilde{\mathbf{e}}_{t+1}^{i}
=
\sum_{v \in \mathcal{T}_{t+1}^{i}}
\tilde{p}_{t+1}^{i}(v) \cdot \mathbf{e}_{v}.
\end{equation}
Then, the current input embedding of suffix token \(i\) is obtained by mixing the \texttt{[MASK]} embedding with the previous-step soft decoded embedding:
\begin{equation}
\tilde{\mathbf{e}}_{t}^{i}
=
(1-\alpha) \cdot \mathbf{e}_{\mathrm{mask}}
+
\alpha \cdot \tilde{\mathbf{e}}_{t+1}^{i},
\end{equation}
where \(\alpha \in [0,1]\) controls the contribution of the previous-step decoding result.

\paragraph{Sparse Input Construction}
After the above process, we can construct the sparse input for the DLM.
Given the prompt \(P\), the current block \(B_t\), and the full suffix \(S_t\), the standard block-wise decoding input can be written as
\[
X_t = [P; B_t; S_t].
\]
Instead of feeding the full suffix into the model, \method constructs a sparse input:
\[
\tilde{X}_t = [P; B_t; \mathcal{R}_t].
\]
The retained suffix tokens keep their original positional encodings~\citep{rope}, allowing the model to still perceive their relative positions in the original sequence.
This is important because the retained middle-region start tokens and the final token are used as structural anchors rather than merely local context tokens.

\paragraph{Early Termination} 
To further enhance efficiency, we also use the early termination technique \citep{chen2026dpad,bao2026learning} that halts generation upon detecting an end-of-sequence (<eos>) token.
This technique performs verification after each decoding block, thereby addressing the computational redundancy caused by fixed-length generation in DLMs.
Its utility is most pronounced in long-sequence settings where the maximum length significantly exceeds the actual generated content.


\section{Experiments}
\subsection{Experimental Settings}
\paragraph{Datasets and Evaluation Metrics}
We evaluate our method on two reasoning datasets (GSM8K~\citep{cobbe2021gsm8k} and MATH~\citep{hendrycksmath2021}) and two code generation datasets (HumanEval~\citep{humaneval} and MBPP~\citep{mbpp}).

We use efficiency and accuracy to evaluate our method.
\textbf{Accuracy} is evaluated using task-specific metrics: mathematical reasoning datasets are measured by flexible-extract and strict-match, while code generation tasks are measured by pass@1.
To assess \textbf{efficiency}, we report the average latency, defined as the end-to-end generation time, as well as Tokens Per Second (TPS) and the generation length ratio ($\bar{\ell}/\ell_{\max}$), which compares the actual generated length to the benchmark setting.

\paragraph{Implementation Details}
All experiments were conducted on NVIDIA A800 40GB GPU.
We evaluate the performance of SSM on three representative dLLMs: LLaDA-8B-Instruct \citep{llada}, LLaDA-1.5 \citep{llada1.5}, and Dream-7B-Base \citep{dream}.
Following~\cite{chen2026dpad}, we use a block size of 32 and a confidence threshold of 0.9 for parallel decoding.
We search for the top-$k$ in the range of [3, 5, 7] and the mixing ratio $\alpha$ is [0.2, 0.3, 0.4].
For the local region size, the search range is [1, 3] with a step size of 0.5 on the GSM8K, MATH, and MBPP datasets, and [3, 5] with a step size of 0.5 on the HumanEval dataset.

\subsection{Baselines}
We use two decoding methods as baselines for comparison.
\textbf{Vanilla} denotes the original top-1 decoding strategy, where only the token with the highest probability is decoded at each step.
\textbf{Parallel} decoding (\textbf{Par.}) \citep{wu2026fastdllm} denotes threshold-based decoding, where all tokens with probabilities above a predefined threshold are decoded.

Since parallel decoding provides substantial speedups over Vanilla decoding, we combine our suffix dropping method with parallel decoding to demonstrate its effectiveness.
\textbf{DPad} \citep{chen2026dpad} defines a local window and applies distance-aware token dropout within the window at the beginning of decoding each block.
\textbf{Streaming} \citep{Streaming-dLLM} preserves all neighboring suffix tokens and the final token to maintain essential contextual and boundary cues, while pruning redundant suffix tokens.
In particular, both \textbf{DPad} and \textbf{Streaming} employ an early termination strategy for fair comparison.
Since our paper focuses exclusively on suffix modeling, the \textbf{Streaming} baseline does not employ its proposed dynamic confidence-aware parallel decoding strategy.

\begin{table*}[t]
\caption{Performance comparison of LLaDA-Instruct and Dream-Base on four benchmarks.}
\label{tab:consolidated-performance}
\centering
\resizebox{\textwidth}{!}{
\begin{tabular}{ll|ccccc|cc|ccccc|cc}
\toprule
\multirow{3}{*}{Benchmark} &
\multirow{3}{*}{Method} &
\multicolumn{7}{c|}{LLaDA-Instruct} &
\multicolumn{7}{c}{Dream-Base} \\
& &
\multicolumn{5}{c}{Efficiency} &
\multicolumn{2}{c|}{Accuracy (\%)} &
\multicolumn{5}{c}{Efficiency} &
\multicolumn{2}{c}{Accuracy (\%)} \\
\cmidrule(lr){3-7} \cmidrule(lr){8-9}
\cmidrule(lr){10-14} \cmidrule(lr){15-16}
& &
\multicolumn{2}{c}{Latency(s)$\downarrow$} &
\multicolumn{2}{c}{TPS$\uparrow$} &
$\bar{\ell}/\ell_{\max}$ &
Flexible$\uparrow$ & Strict$\uparrow$ &
\multicolumn{2}{c}{Latency(s)$\downarrow$} &
\multicolumn{2}{c}{TPS$\uparrow$} &
$\bar{\ell}/\ell_{\max}$ &
Flexible$\uparrow$ & Strict$\uparrow$ \\
\midrule
\multirow{5}{*}{\begin{tabular}[c]{@{}l@{}}GSM8K\\ \textit{4-shot}\end{tabular}}
& Vanilla
& 28.52 & 1.00$\times$ & 8.13 & 1.00$\times$ & 232 / 256 & 78.39 & 37.38
& 22.28 & 1.00$\times$ & 11.44 & 1.00$\times$ & 255 / 256 & \textbf{76.27} & \textbf{75.74} \\
& Par.
& 8.65 & 3.30$\times$ & 26.81 & 3.30$\times$ & 232 / 256 & 78.54 & 38.67
& 14.91 & 1.49$\times$ & 17.10 & 1.49$\times$ & 255 / 256 & 75.51 & 75.51 \\
& DPad+Par.
& 6.68 & 4.27$\times$ & 24.07 & 2.96$\times$ & 160 / 256 & \textbf{79.76} & 64.97
& 5.21 & 4.28$\times$ & 24.29 & 2.12$\times$ & 127 / 256 & 72.78 & 74.75 \\
& Streaming+Par.
& 5.25 & 5.43$\times$ & 21.86 & 2.69$\times$ & 114 / 256 & 78.09 & 74.00
& \textbf{5.12} & \textbf{4.35$\times$} & 24.31 & 2.13$\times$ & 124 / 256 & 75.28 & 75.36 \\
& \cellcolor{tabcolor3}SSM+Par.
& \cellcolor{tabcolor3}\textbf{4.87} & \cellcolor{tabcolor3}\textbf{5.86$\times$} & \cellcolor{tabcolor3}22.63 & \cellcolor{tabcolor3}2.78$\times$ & \cellcolor{tabcolor3}110 / 256 & \cellcolor{tabcolor3}79.38
& \cellcolor{tabcolor3}\textbf{76.95}
& \cellcolor{tabcolor3}5.13 & \cellcolor{tabcolor3}4.34$\times$ & \cellcolor{tabcolor3}24.41 & \cellcolor{tabcolor3}2.13$\times$ & \cellcolor{tabcolor3}125 / 256 & \cellcolor{tabcolor3}75.44 & \cellcolor{tabcolor3}75.44 \\

\midrule
\multirow{5}{*}{\begin{tabular}[c]{@{}l@{}}MATH\\ \textit{4-shot}\end{tabular}}
& Vanilla
& 26.25 & 1.00$\times$ & 9.47 & 1.00$\times$ & 249 / 256 & \textbf{33.66} & 8.42
& 20.84 & 1.00$\times$ & 12.27 & 1.00$\times$ & 255 / 256 & 34.32 & 37.76 \\
& Par. & 10.00 & 2.63$\times$ & 24.84 & 2.62$\times$ & 249 / 256 & 33.48 & 8.76
& 8.76 & 2.38$\times$ & 29.19 & 2.38$\times$ & 255 / 256 & \textbf{35.36} & \textbf{38.62} \\
& DPad+Par. & 9.38 & 2.80$\times$ & 22.48 & 2.37$\times$ & 211 / 256 & 33.54 & 27.98
& 7.53 & 2.77$\times$ & 33.86 & 2.76$\times$ & 255 / 256 & 34.44 & 38.32 \\
& Streaming+Par.
& 7.35 & 3.57$\times$ & 21.41 & 2.26$\times$ & 157 / 256 & 31.38 & 30.68
& 7.75 & 2.69$\times$ & 32.89 & 2.68$\times$ & 255 / 256 & 34.78 & 38.18 \\
& \cellcolor{tabcolor3}SSM+Par.
& \cellcolor{tabcolor3}\textbf{6.11} & \cellcolor{tabcolor3}\textbf{4.30$\times$} & \cellcolor{tabcolor3}21.64 & \cellcolor{tabcolor3}2.29$\times$ & \cellcolor{tabcolor3}132 / 256 & \cellcolor{tabcolor3}32.04
& \cellcolor{tabcolor3}\textbf{31.96}
& \cellcolor{tabcolor3}\textbf{7.37} & \cellcolor{tabcolor3}\textbf{2.83$\times$} & \cellcolor{tabcolor3}34.59 & \cellcolor{tabcolor3}2.82$\times$ & \cellcolor{tabcolor3}255 / 256 & \cellcolor{tabcolor3}34.48 & \cellcolor{tabcolor3}38.00  \\

\midrule
\multirow{5}{*}{\begin{tabular}[c]{@{}l@{}}HumanEval\\ \textit{0-shot}\end{tabular}}
& Vanilla
& 35.09 & 1.00$\times$ & 13.47 & 1.00$\times$ & 473 / 512 & 43.90 & --
& 28.43 & 1.00$\times$ & 17.96 & 1.00$\times$ & 510 / 512 & 51.21 & -- \\
& Par.
& 11.59 & 3.03$\times$ & 40.99 & 3.04$\times$ & 475 / 512 & 43.29 & --
& 14.57 & 1.95$\times$ & 35.06 & 1.95$\times$ & 510 / 512 & 53.04 & -- \\
& DPad+Par.
& 9.23 & 3.80$\times$ & 47.36 & 3.52$\times$ & 437 / 512 & 46.34 & --
& 3.78 & 7.52$\times$ & 56.54 & 3.15$\times$ & 214 / 512 & 52.43 & -- \\
& Streaming+Par.
& 5.41 & 6.49$\times$ & 54.32 & 4.03$\times$ & 293 / 512 & 46.34 & --
& 3.83 & 7.42$\times$ & 54.99 & 3.06$\times$ & 210 / 512 & 52.43 & -- \\
& \cellcolor{tabcolor3}SSM+Par.
& \cellcolor{tabcolor3}\textbf{3.35} & \cellcolor{tabcolor3}\textbf{10.47$\times$} & \cellcolor{tabcolor3}65.76 & \cellcolor{tabcolor3}4.88$\times$ & \cellcolor{tabcolor3}220 / 512 & \cellcolor{tabcolor3}\textbf{48.17}
& \cellcolor{tabcolor3}--
& \cellcolor{tabcolor3}\textbf{3.40} & \cellcolor{tabcolor3}\textbf{8.36$\times$} & \cellcolor{tabcolor3}55.72 & \cellcolor{tabcolor3}3.10$\times$ & \cellcolor{tabcolor3}189 / 512 & \cellcolor{tabcolor3}\textbf{53.04} & \cellcolor{tabcolor3}-- \\

\midrule
\multirow{5}{*}{\begin{tabular}[c]{@{}l@{}}MBPP\\ \textit{3-shot}\end{tabular}}
& Vanilla
& 62.82 & 1.00$\times$ & 4.76 & 1.00$\times$ & 299 / 512 & 15.00 & --
& 49.07 & 1.00$\times$ & 10.43 & 1.00$\times$ & 512 / 512 & 52.40 & -- \\
& Par. 
& 14.43 & 4.35$\times$ & 20.74 & 4.36$\times$ & 299 / 512 & 15.00 & --
& 12.35 & 3.97$\times$ & 41.45 & 3.97$\times$ & 512 / 512 & \textbf{56.20} & -- \\
& DPad+Par. 
& 6.03 & 10.42$\times$ & 18.22 & 3.83$\times$ & 110 / 512 & 39.40 & --
& 9.85 & 4.98$\times$ & 51.96 & 4.98$\times$ & 512 / 512 & 54.80 & -- \\
& Streaming+Par.
& 6.17 & 10.18$\times$ & 16.34 & 3.43$\times$ & 100 / 512 & 42.40 & --
& 9.66 & 5.08$\times$ & 52.94 & 5.08$\times$ & 512 / 512 & 54.00 & -- \\
& \cellcolor{tabcolor3}SSM+Par.
& \cellcolor{tabcolor3}\textbf{2.68} & \cellcolor{tabcolor3}\textbf{23.44$\times$} & \cellcolor{tabcolor3}19.44 & \cellcolor{tabcolor3}4.08$\times$ & \cellcolor{tabcolor3}52 / 512 & \cellcolor{tabcolor3}\textbf{45.20}
& \cellcolor{tabcolor3}-- & \cellcolor{tabcolor3}\textbf{8.99} & \cellcolor{tabcolor3} \textbf{5.46$\times$} & \cellcolor{tabcolor3}56.89 & \cellcolor{tabcolor3}5.45$\times$ & \cellcolor{tabcolor3}512 / 512 & \cellcolor{tabcolor3}55.60 & \cellcolor{tabcolor3}-- \\

\bottomrule
\end{tabular}
}
\end{table*}

\subsection{Main Results}
Table~\ref{tab:consolidated-performance} compares Vanilla decoding, parallel decoding, DPad+Par., Streaming+Par., and our method combined with Par. on four benchmarks using LLaDA-Instruct and Dream-Base.
We also report the results using LLaDA-1.5 in Table~\ref{tab:llada1.5}.
Table~\ref{tab:consolidated-performance} shows that SSM outperforms all baseline methods in most cases, achieving the lowest latency.
Notably, compared with the strongest baseline method, Streaming+Par., our method achieves a 2.3× speedup on the MBPP dataset using LLaDA-Instruct.
Compared with Par. alone, our method further reduces latency by 43.7\%, 38.9\%, 71.1\%, and 81.4\% on LLaDA-Instruct.
This also demonstrates that our method can be further combined with decoding strategies for acceleration.

\paragraph{Efficiency.}
The efficiency gains come from two sources.
First, \method reduces per-step suffix computation by assigning different token budgets to different suffix regions.
Second, \method often produces more concise generations, as indicated by smaller \(\bar{\ell}/\ell_{\max}\) values.
This reduction is not attributed to early termination; instead, it arises because SSM exposes each block to fewer suffix tokens, reducing redundant contextual information and leading to more compact continuations.
The reduction in output length is particularly pronounced in code generation tasks: on MBPP with LLaDA-Instruct, \(\bar{\ell}/\ell_{\max}\) decreases from 299/512 to 52/512.
As a result, \method achieves substantially larger latency reductions on HumanEval and MBPP than on GSM8K and MATH, since longer generations involve more suffix computation and thus provide greater opportunities for suffix pruning. In contrast, multi-shot mathematical reasoning benchmarks typically feature longer input prompts but shorter outputs, limiting the achievable speedup from our suffix pruning strategy.
Beyond this, the improvement of our method on Dream is less pronounced than that on LLaDA-Instruct.
We attribute this to differences in their training strategies: LLaDA is trained from scratch, whereas Dream is initialized from an autoregressive model.
As a result, LLaDA’s native DLM architecture may be more sensitive to suffix tokens, making it more compatible with our method.

We find that latency and TPS are not always aligned.
This reflects an inherent limitation of TPS as an efficiency metric~\citep{qian2026d3llm,chen2026dpad}: when the generation length is substantially shortened, reduced GPU utilization can mechanically lower TPS, even though the end-to-end latency is improved.
This motivates the community to seek more appropriate throughput metrics for evaluating performance in future work.

\paragraph{Accuracy.}
In terms of accuracy, \method maintains competitive performance and often improves over strong baselines.
Our improvements are more significant on code tasks and on the strict-match metric for mathematical tasks.
Notably, strict-match requires not only generating the correct answer but also following the specific reasoning format demonstrated in the examples, thus reflecting the model’s in-context learning ability to some extent.
By retaining a small number of suffix tokens, our method encourages the model to produce concise and well-formatted outputs as early as possible, thereby avoiding erroneous intermediate steps.
These gains also suggest that suffix dropout does not necessarily trade accuracy for speed.


\begin{table*}[t]
\centering
\caption{Ablation study of LLaDA-Instruct on two datasets.}
\label{tab:ablation}
\resizebox{\textwidth}{!}{
\begin{tabular}{l|ccccc|cc|ccccc|c}
\toprule
\multirow{3}{*}{Method} &
\multicolumn{7}{c|}{GSM8K} &
\multicolumn{6}{c}{HumanEval} \\
&
\multicolumn{5}{c}{Efficiency} &
\multicolumn{2}{c|}{Accuracy (\%)} &
\multicolumn{5}{c}{Efficiency} &
Accuracy (\%) \\
\cmidrule(lr){2-6} \cmidrule(lr){7-8}
\cmidrule(lr){9-13} \cmidrule(lr){14-14}
&
\multicolumn{2}{c}{Latency(s)$\downarrow$} &
\multicolumn{2}{c}{TPS$\uparrow$} &
$\bar{\ell}/\ell_{\max}$ &
Flexible$\uparrow$ & Strict$\uparrow$ &
\multicolumn{2}{c}{Latency(s)$\downarrow$} &
\multicolumn{2}{c}{TPS$\uparrow$} &
$\bar{\ell}/\ell_{\max}$ &
pass@1$\uparrow$ \\
\midrule
\textbf{Vanilla}
& 28.52 & 1.00$\times$ & 8.13 & 1.00$\times$ & 232/256 & 78.39 & 37.38 & 35.09 & 1.00$\times$ & 13.47 & 1.00$\times$ & 473/512 & 43.90 \\
\cellcolor{tabcolor3}\textbf{SSM+Par. (\textbf{\textit{Ours}}) }
& \cellcolor{tabcolor3}\textbf{4.87} & \cellcolor{tabcolor3}\textbf{5.86$\times$} & \cellcolor{tabcolor3}\textbf{22.63} & \cellcolor{tabcolor3}\textbf{2.78$\times$} & \cellcolor{tabcolor3}\textbf{110/256} & \cellcolor{tabcolor3}\textbf{79.38}
& \cellcolor{tabcolor3}\textbf{76.95} & \cellcolor{tabcolor3}\textbf{3.35} & \cellcolor{tabcolor3}\textbf{10.47$\times$} & \cellcolor{tabcolor3}\textbf{65.76} & \cellcolor{tabcolor3}\textbf{4.88$\times$} & \cellcolor{tabcolor3}\textbf{220/512} & \cellcolor{tabcolor3}\textbf{48.17}  \\
\midrule
\multicolumn{14}{l}{\it{Local Region:}}\\
\textbf{w/o Local Region} & 1.96 & 14.55$\times$ & 14.36 & 1.77$\times$ & 28/256 & 33.51 & 25.32 & 0.51& 68.80$\times$ & 13.26 & 1.02$\times$ & 6 / 512  & 15.24\\
\midrule
\multicolumn{14}{l}{\it{Middle Region:}}\\
\textbf{w/o Start Token} & 4.95 & 5.76$\times$ & 22.82 & 2.81$\times$ & 113/256 & 78.32 & 75.59 & 3.59 & 9.77$\times$ & 65.25 & 4.84$\times$& 234/512 & 43.29\\
\textbf{\makecell{Replace the start token\\ with the end token}}& 4.79 & 5.95$\times$ & 22.71 & 2.79$\times$ & 109/256 & 78.62 & 75.82 & 3.30 & 10.63$\times$ & 65.14 &4.84$\times$ & 214/512 & 46.34\\
\textbf{Start+End Tokens} & 4.83 & 5.90$\times$ & 22.57 & 2.78$\times$ & 109/256 & 79.98 & 77.33 & 3.20 & 10.97$\times$ & 67.03 & 4.98$\times$ & 214/512 & 43.29 \\
\midrule
\multicolumn{14}{l}{\it{Tail Region:}}\\
\textbf{Start Token Only} & 4.95 & 5.76$\times$ & 22.62 & 2.78$\times$ & 112/256 & 78.85 & 76.19 & 3.30 & 10.63$\times$ & 65.05 & 4.83$\times$ & 215/512 & 43.29\\
\textbf{End Token Only} & 4.87 & 5.86$\times$ & 22.65 & 2.79$\times$ & 110/256 & 79.61 & 76.80 & 3.29 & 10.67$\times$ &65.58 & 4.87$\times$ & 216/512 & 43.29\\
\textbf{Start+Mid+End Tokens} & 4.83 & 5.90$\times$ & 22.83 & 2.81$\times$ & 110/256 & 78.62 & 76.12 & 3.42 & 10.26$\times$ & 65.56 & 4.87$\times$ & 224/512 & 45.73\\
\midrule
\multicolumn{14}{l}{\it{Soft Embedding:}}\\
\textbf{w/o Soft Embedding} & 5.06 &5.64$\times$ & 22.09 & 2.72$\times$& 111/256 & 77.41 & 73.54 & 4.26 & 8.24$\times$ & 59.54 & 4.42$\times$ & 253/512 & 48.17\\
\midrule
\multicolumn{14}{l}{\it{Early Termination:}}\\
\textbf{w/o Early Termination} & 5.34 & 5.34$\times$ & 20.61 & 2.54$\times$& 110/256 & 79.38 & 76.95 & 3.96 & 8.86$\times$ & 55.94 & 4.15$\times$ & 220/512 & 48.17\\
\bottomrule
\end{tabular}
}
\end{table*}

\subsection{Ablation Study}
Table~\ref{tab:ablation} studies the contribution of each component.
All of the proposed components contributed to the performance improvement.

Removing the \textit{local region} leads to severe performance degradation, and the average generation length also becomes extremely short.
This suggests that neighboring suffix tokens provide essential fine-grained context and prevent premature or incomplete generation.
Thus, the local region should be densely retained.
For the \textit{middle region}, removing start tokens not only hurts accuracy on both datasets but also increases latency.
This indicates that retaining start tokens in the middle region is necessary for preserving the textual skeleton.
Replacing start tokens with end tokens slightly reduces latency but degrades accuracy. This may be because end tokens are better at locating the end boundary rather than preserving block-level structural cues.
Notably, additionally retaining end tokens in the middle region does not lead to consistent performance improvements.
For the \textit{tail region}, retaining both the start and end tokens achieves more stable performance on the two datasets.
Retaining only the start token or only the end token leads to comparable performance, with the end token yielding better results on GSM8K.
In addition, introducing extra middle tokens does not bring further performance gains.
This suggests that retaining the start and end tokens in the final block is sufficient.
Notably, since the tail region contains only one block, removing or adding a single token does not lead to a noticeable change in latency.

\textit{Early termination} fills the remaining blocks with EOS tokens once an EOS token is detected; therefore, it only eliminates the time required to generate subsequent blocks without changing the generation length.
\textit{Soft suffix embedding mixing} reduces latency and improves performance on GSM8K, while substantially shortening the generation length and reducing latency on HumanEval.
This shows that incorporating previous-step decoding results into current suffix representations helps preserve evolving denoising information and encourages more efficient convergence.

\begin{figure}[t]
  \centering
  \includegraphics[width=\textwidth]{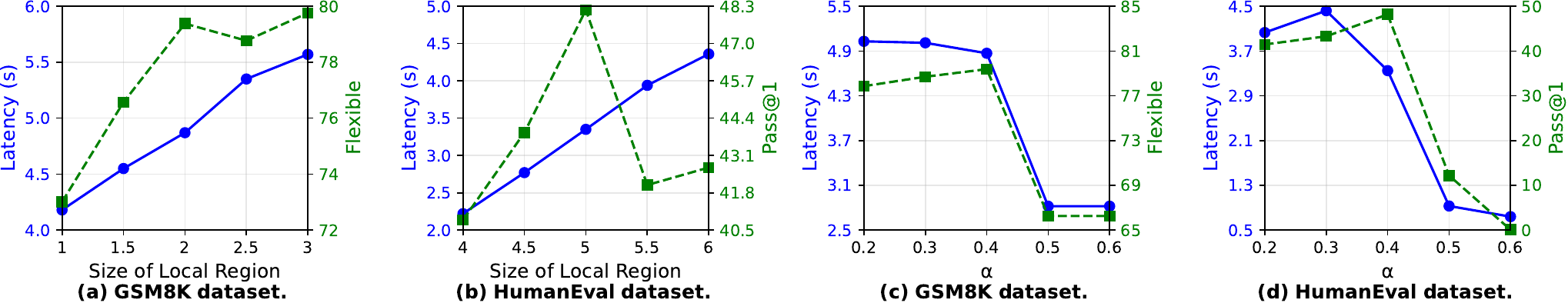}
  \caption{Effects of hyper-parameters on two datasets.}
  \label{fig:hyper}
\end{figure}

\subsection{Effects of Hyper-Parameters}
\paragraph{Effects of the size of local region.}
We explore the effects of local region size in Figures~\ref{fig:hyper}(a) and \ref{fig:hyper}(b).
Increasing the local region size leads to higher latency on both datasets, since more retained suffix tokens introduce additional computation.
On GSM8K, performance generally improves with a larger local region.
This indicates that, for the GSM8K dataset, the information provided by the local region is highly beneficial to the overall performance.
However, HumanEval is more sensitive to changes in the hyperparameter. It achieves the highest accuracy when the local region size is set to 5.
This result indicates that different tasks require different local region sizes to balance performance and computational cost.

\paragraph{Effects of $\alpha$.}
We then explore the effects of $\alpha$ in Figures~\ref{fig:hyper}(c) and \ref{fig:hyper}(d).
When $\alpha$ is set to 0.2, 0.3, or 0.4, both latency and accuracy remain relatively stable on the two datasets.
The best accuracy on both datasets is achieved at $\alpha=0.4$.
However, when $\alpha$ is further increased to 0.5 or 0.6, accuracy drops sharply.
This is likely because predictions from early timesteps are still unreliable; assigning them overly large weights may corrupt suffix token representations and lead to representation collapse.
Overall, $\alpha=0.4$ provides the best accuracy while maintaining low latency on both datasets.
This further indicates that the setting of $\alpha$ generalizes well across different tasks.

\subsection{Comparison with Other Parallel Decoding Methods}
We further compare our method with a margin-based parallel decoding strategy~\citep{kim2025train} to verify its effectiveness.
In this strategy, the decision to decode a token depends on whether the probability gap between the top two predictions exceeds a predefined threshold.

\begin{wraptable}{r}{0.55\textwidth}
\caption{Performance of margin-based parallel decoding methods on two datasets using LLaDA-Instruct.}
\label{tab:margin-comparison}
\centering
\resizebox{0.55\textwidth}{!}{
\begin{tabular}{clccc}
\toprule
Benchmark & Method & Acc. (Flex. / Str.) & Lat.(s) / TPS & Speedup \\
\midrule
\multirow{4}{*}{\shortstack{GSM8K\\\textit{4-shot}}}
& Margin
& 78.54 / 38.67
& 8.52 / \textbf{27.20}
& 1.00$\times$ \\

& DPad+Margin
& 79.00 / 65.88
& 7.65 / 20.78
& 1.11$\times$ \\

& Streaming+Margin
& 77.86 / 74.07
& 6.07 / 19.00
& 1.40$\times$ \\

& SSM+Margin
& \textbf{79.61 / 76.65}
& \textbf{5.57} / 19.83
& \textbf{1.53$\times$} \\

\midrule

\multirow{4}{*}{\shortstack{HumanEval\\\textit{0-shot}}}
& Margin
& 43.29 / --
& 11.54 / 41.16
& 1.00$\times$ \\

& DPad+Margin
& 42.68 / --
& 10.65 / 40.89
& 1.08$\times$ \\

& Streaming+Margin
& 42.68 / --
& 5.74 / 49.41
& 2.01$\times$ \\

& SSM+Margin
& \textbf{43.90} / --
& \textbf{3.47 / 60.73}
& \textbf{3.33$\times$} \\

\bottomrule
\end{tabular}
}
\end{wraptable}

Table \ref{tab:margin-comparison} shows that our method achieves the best accuracy and latency compared with the three baselines.
Notably, our method further accelerates the margin-based decoding strategy, achieving a \(3.33\times\) speedup on HumanEval.
These results further demonstrate the effectiveness and generalizability of our method.
Meanwhile, they show that suffix dropping methods are orthogonal to parallel decoding strategies and can be combined with them to further improve both speed and performance.


\begin{wrapfigure}{r}{0.65\columnwidth}
  \centering
  \includegraphics[width=0.65\columnwidth]{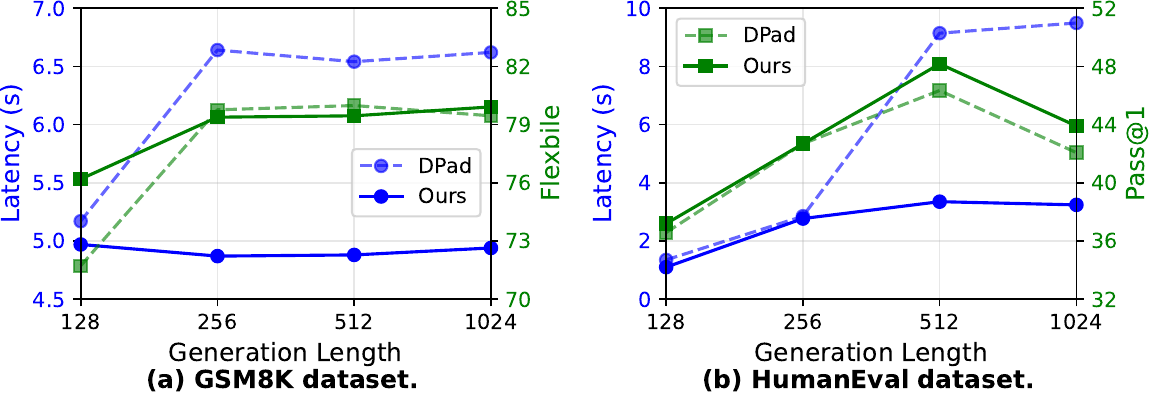}
  \caption{Effects of generation length on two datasets.}
  \label{fig:length}
\end{wrapfigure}

\subsection{Effects of Generation Length}
We further compare our method with DPad under different generation lengths on two datasets.

Our method achieves lower latency across all generation lengths, with the advantage becoming more pronounced as the generation length increases in Figure~\ref{fig:length}.
For example, on HumanEval with a generation length of 1024, our method is nearly \(3\times\) faster than DPad.
This indicates that our method can accelerate inference across various generation lengths.
The overall trend remains that shorter generation lengths lead to lower latency. Our early termination strategy can reduce the overhead caused by overly long generations to some extent.
In terms of accuracy, our method achieves performance comparable to DPad on GSM8K and better performance on HumanEval.
Moreover, a longer generation length does not necessarily lead to better performance.

\subsection{Performance on Long-sequence Generation}
To further evaluate the efficiency of SSM in long-sequence generation, we follow \citet{chen2026dpad} and report the results on GSM8K using LLaDA-1.5 with a maximum generation length of 1024 tokens. As shown in Table~\ref{tab:llada15-gsm8k}, we combine SSM with parallel decoding (+Par.)~\citep{wu2026fastdllm} and prefix caching (+PC.)~\citep{wu2026fastdllm} to examine its compatibility with decoding-based and cache-based acceleration techniques. We compare SSM with DPad under the same acceleration settings.

\begin{wraptable}{r}{0.48\textwidth}
    \centering
    \caption{Performance on GSM8K with LLaDA-1.5 (1024 tokens, 1-shot).}
    \label{tab:llada15-gsm8k}
    \resizebox{0.48\textwidth}{!}{
    \begin{tabular}{lcccc}
    \toprule
    \textbf{Method} &
    \textbf{\begin{tabular}[c]{@{}c@{}}Acc.\\(Flex. / Str.)\end{tabular}} &
    \textbf{Lat.(s) / TPS} &
    \textbf{Speedup} \\
    \midrule

    Par.
    & 78.77 / 49.43
    & 11.98 / 16.43
    & 1.00$\times$ \\

    Par.+DPad
    & \textbf{80.89} / 70.36
    & 2.74 / 52.61
    & 4.37$\times$ \\
    Par.+SSM
    & 79.91 / \textbf{74.37}
    & \textbf{2.18} / 46.94
    & \textbf{5.49$\times$} \\
    \midrule
    Par.+PC.
    & 79.76 / 51.63
    & 10.85 / 18.09
    & 1.00$\times$ \\
    Par.+PC.+DPad
    & 80.67 / 72.10
    & 2.14 / \textbf{65.97}
    & 5.07$\times$ \\
    Par.+PC.+SSM
    & \textbf{80.74} / \textbf{75.13}
    & \textbf{1.76} / 58.02
    & \textbf{6.16$\times$} \\

    \midrule

    Vanilla
    & 78.17 / 48.98
    & 128.16 / 1.54
    & 1.00$\times$ \\

    \multicolumn{3}{l}{
    \textbf{Overall} (Par.+PC.+SSM vs. Vanilla)
    }
    & \textbf{72.81$\times$} \\

    \bottomrule
    \end{tabular}
    }
\end{wraptable}

SSM consistently achieves lower latency than DPad and obtains better accuracy in most cases across different combinations of acceleration techniques, further confirming the effectiveness of our method.
It is worth noting that long-sequence generation scenarios involve longer suffixes, where suffix dropout methods can achieve more substantial improvements.

In addition, SSM can be combined with KV cache and parallel decoding for further speedup, confirming that it is orthogonal to these techniques.
In addition, both DPad and our method lead to accuracy improvements.
We attribute this to the fact that suffix dropping often shortens the generation length, which helps reduce the likelihood of generating erroneous continuations.
Overall, compared with Vanilla top-1 decoding, our method combined with Par. and PC. achieves a 72.81$\times$ speedup.

\section{Conclusion}
In this paper, we propose SSM, a training-free framework for accelerating DLM inference through structured suffix modeling.
Our method exploits the structural heterogeneity of suffix tokens by assigning different token budgets to local, middle, and tail regions, thereby preserving essential contextual, structural, and boundary cues while reducing redundant suffix computation.
We further introduce soft suffix embeddings to incorporate previous step decoding information into current suffix representations, preserving evolving denoising signals across generation steps.
Experiments on mathematical reasoning and code generation benchmarks show that \method effectively reduces inference latency while maintaining competitive accuracy.
Moreover, \method is orthogonal to existing acceleration techniques such as parallel decoding and KV caching, enabling further efficiency gains when combined with them.
Notably, in long-sequence inference, our method achieves up to a \(72.81\times\) speedup when combined with other acceleration techniques.



\bibliographystyle{apalike}  
\small
\bibliography{custom}


\appendix

\section{Limitations}
First, the efficiency gain of our method depends on the suffix length. When the suffix constitutes only a small proportion of the sequence, e.g., with a long prefix or a short generation length, the efficiency improvement is relatively limited.
Second, our method may cause performance degradation on some datasets, making it less suitable for scenarios where accuracy is highly critical.
Finally, we have not evaluated our method on models larger than 10B parameters.

\section{Results on LLaDA-1.5}
Table~\ref{tab:llada1.5} reports the performance of LLaDA-1.5 across four benchmarks, further validating the effectiveness of the SSM strategy on advanced diffusion language models.

\textbf{Efficiency.} In terms of inference efficiency, SSM+Par. demonstrates superior and stable acceleration across all benchmarks.
Compared to DPad and Streaming, SSM consistently maintains the lowest latency.
This efficiency is further evidenced by the reduction in average generated sequence length ($\bar{\ell}/\ell_{\max}$), with values reaching as low as 73/512 on HumanEval and 49/512 on MBPP.
These results indicate that the SSM strategy induces the model to converge more rapidly to valid outputs, thereby reducing computational redundancy by minimizing the generation of unnecessary long sequences.

\textbf{Accuracy.} SSM+Par. can reliably maintain task performance and sometimes further improve it.
On GSM8K, it achieves the highest Flexible and Strict accuracy scores.
Notably, compared to Vanilla, our method improves the Strict accuracy from 61.87\% to 80.14\%, effectively bridging the performance gap typically caused by non-standard output formats.
Compared to the results on LLaDA-8B-Instruct, the performance gains on LLaDA-1.5 are even more pronounced, suggesting that the effectiveness of SSM can be further amplified by stronger base models.

\section{Comparison under Vanilla Top-1 Decoding}
We further compare our method with the baselines under vanilla Top-1 decoding in Table~\ref{tab:vanilla}.
Results show that SSM achieves the best latency and accuracy in most cases.
These results demonstrate that our method remains effective under Top-1 decoding, indicating its compatibility with different decoding strategies.

\begin{table*}[t]
\caption{Performance of LLaDA-1.5 on four benchmarks.}
\label{tab:llada1.5}
\centering
\resizebox{\textwidth}{!}{
\begin{tabular}{ll|ccccc|cc}
\toprule
\multirow{3}{*}{Benchmark} &
\multirow{3}{*}{Method} &
\multicolumn{7}{c}{LLaDA-1.5} \\
& &
\multicolumn{5}{c|}{Efficiency} &
\multicolumn{2}{c}{Accuracy (\%)} \\
\cmidrule(lr){3-7} \cmidrule(lr){8-9}
& &
Latency(s)$\downarrow$ &
Speedup$\uparrow$ &
TPS$\uparrow$ &
Speedup$\uparrow$ &
$\bar{\ell}/\ell_{\max}$ &
Flexible$\uparrow$ &
Strict$\uparrow$ \\
\midrule

\multirow{5}{*}{\shortstack{GSM8K\\\textit{4-shot}}}
& Vanilla & 27.70 & 1.00$\times$ & 7.74 & 1.00$\times$ & 214 / 256 & 80.59 & 61.87 \\
& Par. & 8.28 & 3.35$\times$ & 25.91 & 3.35$\times$ & 214 / 256 & 80.82 & 62.62 \\
& DPad+Par. & 6.37 & 4.35$\times$ & 24.67 & 3.19$\times$ & 157 / 256 & 80.89 & 78.92 \\
& Streaming+Par. & 5.26 & 5.27$\times$ & 23.07 & 2.98$\times$ & 121 / 256 & 79.15 & 78.09 \\
& SSM+Par. & \textbf{4.96} & \textbf{5.58$\times$} & 23.96 & 3.10$\times$ & 110 / 256 & \textbf{81.50} & \textbf{80.14} \\

\midrule

\multirow{5}{*}{\shortstack{MATH\\\textit{4-shot}}}
& Vanilla & 25.70 & 1.00$\times$ & 8.47 & 1.00$\times$ & 217 / 256 & 33.62 & 32.72 \\
& Par. & 9.89 & 2.60$\times$ & 21.99 & 2.60$\times$ & 217 / 256 & \textbf{33.72} & 32.92 \\
& DPad+Par. & 8.74 & 2.94$\times$ & 22.32 & 2.64$\times$ & 195 / 256 & 33.08 & \textbf{35.96} \\
& Streaming+Par. & 7.13 & 3.60$\times$ & 21.82 & 2.58$\times$ & 155 / 256 & 32.54 & 35.74 \\
& SSM+Par. & \textbf{5.79} & \textbf{4.44$\times$} & 22.34 & 2.64$\times$ & 129 / 256 & 32.24 & 35.86 \\

\midrule

\multirow{5}{*}{\shortstack{HumanEval\\\textit{0-shot}}}
& Vanilla & 34.81 & 1.00$\times$ & 3.15 & 1.00$\times$ & 109 / 512 & 40.85 & -- \\
& Par. & 11.54 & 3.02$\times$ & 9.47 & 3.01$\times$ & 109 / 512 & 39.63 & -- \\
& DPad+Par. & 5.19 & 6.71$\times$ & 16.19 & 5.14$\times$ & 84 / 512 & 39.63 & -- \\
& Streaming+Par. & 8.27 & 4.21$\times$ & 9.91 & 3.15$\times$ & 141 / 512 & 39.63 & -- \\
& SSM+Par. & \textbf{4.77} & \textbf{7.30$\times$} & 15.38 & 4.88$\times$ & 73 / 512 & \textbf{45.73} & -- \\

\midrule

\multirow{5}{*}{\shortstack{MBPP\\\textit{3-shot}}}
& Vanilla & 62.65 & 1.00$\times$ & 1.01 & 1.00$\times$ & 63 / 512 & 38.20 & -- \\
& Par. & 5.66 & 11.07$\times$ & 11.22 & 11.11$\times$ & 63 / 512 & 38.60 & -- \\
& DPad+Par. & 4.48 & 13.98$\times$ & 14.57 & 14.43$\times$ & 65 / 512 & 41.60 & -- \\
& Streaming+Par. & 3.79 & 16.53$\times$ & 21.54 & 21.33$\times$ & 81 / 512 & 41.40 & -- \\
& SSM+Par. & \textbf{2.66} & \textbf{23.55$\times$} & 18.50 & 18.32$\times$ & 49 / 512 & \textbf{42.40} & -- \\

\bottomrule
\end{tabular}
}
\end{table*}

\begin{table*}[t]
\caption{Performance comparison of LLaDA-Instruct using the Top-1 decoding strategy on three benchmarks.}
\label{tab:vanilla}
\centering
\resizebox{\textwidth}{!}{
\begin{tabular}{ll|ccccc|cc}
\toprule
\multirow{3}{*}{Benchmark} &
\multirow{3}{*}{Method} &
\multicolumn{7}{c}{LLaDA-Instruct} \\
& &
\multicolumn{5}{c}{Efficiency} &
\multicolumn{2}{c}{Accuracy (\%)}  \\
\cmidrule(lr){3-7} \cmidrule(lr){8-9}
& &
\multicolumn{2}{c}{Latency(s)$\downarrow$} &
\multicolumn{2}{c}{TPS$\uparrow$} &
$\bar{\ell}/\ell_{\max}$ &
Flexible$\uparrow$ & Strict$\uparrow$  \\
\midrule
\multirow{4}{*}{\begin{tabular}[c]{@{}l@{}}GSM8K\\ \textit{4-shot}\end{tabular}}
& Vanilla
& 28.52 & 1.00$\times$ & 8.13 & 1.00$\times$ & 232 / 256 & 78.39 & 37.38 \\
& DPad+Van. & 18.80 & 1.52$\times$ & 8.54 & 1.05$\times$ & 160 / 256 & 78.54 & 63.84 \\
& Streaming+Van. & 13.73 & 2.08$\times$ & 8.35 & 1.03$\times$ & 114 / 256 & 78.32 & 73.69 \\
& \cellcolor{tabcolor3}SSM+Van. & \cellcolor{tabcolor3}\textbf{13.51} & \cellcolor{tabcolor3}\textbf{2.11$\times$} & \cellcolor{tabcolor3}8.14 & \cellcolor{tabcolor3}1.00$\times$ & \cellcolor{tabcolor3}110 / 256 & \cellcolor{tabcolor3}\textbf{79.68}
& \cellcolor{tabcolor3}\textbf{76.65} \\
\midrule
\multirow{4}{*}{\begin{tabular}[c]{@{}l@{}}HumanEval\\ \textit{0-shot}\end{tabular}}
& Vanilla
& 34.67 & 1.00$\times$ & 13.64 & 1.00$\times$ & 473 / 512 & 43.90 & -- \\
& DPad+Van.
& 27.13 & 1.52$\times$ & 16.11 & 1.05$\times$ & 437 / 512 & \textbf{47.56} & -- \\
& Streaming+Van. & 16.78 & 1.28$\times$ & 17.05 & 1.18$\times$ & 286 / 512 & 45.73 & -- \\
& \cellcolor{tabcolor3}SSM+Van. & \cellcolor{tabcolor3}\textbf{12.64} & \cellcolor{tabcolor3}\textbf{2.74$\times$} & \cellcolor{tabcolor3}16.67 & \cellcolor{tabcolor3}1.22$\times$ & \cellcolor{tabcolor3}210 / 512 & \cellcolor{tabcolor3}46.34
& \cellcolor{tabcolor3}-- \\

\midrule
\multirow{4}{*}{\begin{tabular}[c]{@{}l@{}}MBPP\\ \textit{3-shot}\end{tabular}}
& Vanilla
& 62.11 & 1.00$\times$ & 4.82 & 1.00$\times$ & 299 / 512 & 15.00 & -- \\
& DPad+Van. 
& 16.25 & 3.82$\times$ & 6.70 & 1.39$\times$ & 108 / 512 & 40.40 & -- \\
& Streaming+Van.
& 12.42 & 5.00$\times$ & 7.17 & 1.49$\times$ & 89 / 512 & 41.60 & -- \\
& \cellcolor{tabcolor3}SSM+Van.
& \cellcolor{tabcolor3}\textbf{7.40} & \cellcolor{tabcolor3}\textbf{8.39$\times$} & \cellcolor{tabcolor3}7.00 & \cellcolor{tabcolor3}1.45$\times$ & \cellcolor{tabcolor3}51 / 512 & \cellcolor{tabcolor3}\textbf{43.00}
& \cellcolor{tabcolor3}--  \\

\bottomrule
\end{tabular}
}
\end{table*}

\newpage
\section*{NeurIPS Paper Checklist}

\begin{enumerate}

\item {\bf Claims}
    \item[] Question: Do the main claims made in the abstract and introduction accurately reflect the paper's contributions and scope?
    \item[] Answer: \answerYes{} 
    \item[] Justification: We have described our claims and contributions clearly in the abstract and introduction.

\item {\bf Limitations}
    \item[] Question: Does the paper discuss the limitations of the work performed by the authors?
    \item[] Answer: \answerYes{} 
    \item[] Justification: We provide a Limitaion section in Appendix.

\item {\bf Theory assumptions and proofs}
    \item[] Question: For each theoretical result, does the paper provide the full set of assumptions and a complete (and correct) proof?
    \item[] Answer: \answerNA{} 
    \item[] Justification: We prove our claims with extensive empirical results, instead of theoretical proofs.

    \item {\bf Experimental result reproducibility}
    \item[] Question: Does the paper fully disclose all the information needed to reproduce the main experimental results of the paper to the extent that it affects the main claims and/or conclusions of the paper (regardless of whether the code and data are provided or not)?
    \item[] Answer: \answerYes{} 
    \item[] Justification: We provided the hyperparameters to reproduce the main experimental results in the Experimental Settings.

\item {\bf Open access to data and code}
    \item[] Question: Does the paper provide open access to the data and code, with sufficient instructions to faithfully reproduce the main experimental results, as described in supplemental material?
    \item[] Answer: \answerYes{}{} 
    \item[] Justification: We provide data and code in Github.

\item {\bf Experimental setting/details}
    \item[] Question: Does the paper specify all the training and test details (e.g., data splits, hyperparameters, how they were chosen, type of optimizer) necessary to understand the results?
    \item[] Answer: \answerYes{} 
    \item[] Justification: We have provided these details in the Experimental Settings.

\item {\bf Experiment statistical significance}
    \item[] Question: Does the paper report error bars suitably and correctly defined or other appropriate information about the statistical significance of the experiments?
    \item[] Answer: \answerNo{} 
    \item[] Justification: The proposed method is training-free and evaluated under a fixed inference-only setting, so there is no variance from random initialization or training runs. Therefore, we do not report error bars or statistical significance tests.

\item {\bf Experiments compute resources}
    \item[] Question: For each experiment, does the paper provide sufficient information on the computer resources (type of compute workers, memory, time of execution) needed to reproduce the experiments?
    \item[] Answer: \answerYes{} 
    \item[] Justification: All experiments are conducted on 4 NVIDIA A800 GPUs, as described in the Experimental Settings.
    
\item {\bf Code of ethics}
    \item[] Question: Does the research conducted in the paper conform, in every respect, with the NeurIPS Code of Ethics \url{https://neurips.cc/public/EthicsGuidelines}?
    \item[] Answer: \answerYes{} 
    \item[] Justification: Our research conforms to the NeurIPS Code of Ethics.

\item {\bf Broader impacts}
    \item[] Question: Does the paper discuss both potential positive societal impacts and negative societal impacts of the work performed?
    \item[] Answer: \answerNo{} 
    \item[] Justification: Our method focuses on accelerating diffusion language models through suffix-based optimization. Beyond this technical contribution, we do not foresee any direct societal impact.
    
\item {\bf Safeguards}
    \item[] Question: Does the paper describe safeguards that have been put in place for responsible release of data or models that have a high risk for misuse (e.g., pre-trained language models, image generators, or scraped datasets)?
    \item[] Answer: \answerNA{} 
    \item[] Justification: Our paper does not have such risks.

\item {\bf Licenses for existing assets}
    \item[] Question: Are the creators or original owners of assets (e.g., code, data, models), used in the paper, properly credited and are the license and terms of use explicitly mentioned and properly respected?
    \item[] Answer: \answerYes{} 
    \item[] Justification: All of the creators or original owners of assets used in the paper are cited properly.

\item {\bf New assets}
    \item[] Question: Are new assets introduced in the paper well documented and is the documentation provided alongside the assets?
    \item[] Answer: \answerYes{} 
    \item[] Justification: We will release the data and code, including documentation, after the paper is accepted.

\item {\bf Crowdsourcing and research with human subjects}
    \item[] Question: For crowdsourcing experiments and research with human subjects, does the paper include the full text of instructions given to participants and screenshots, if applicable, as well as details about compensation (if any)? 
    \item[] Answer: \answerNA{} 
    \item[] Justification: The paper does not involve crowdsourcing nor research with human subjects.

\item {\bf Institutional review board (IRB) approvals or equivalent for research with human subjects}
    \item[] Question: Does the paper describe potential risks incurred by study participants, whether such risks were disclosed to the subjects, and whether Institutional Review Board (IRB) approvals (or an equivalent approval/review based on the requirements of your country or institution) were obtained?
    \item[] Answer: \answerNA{} 
    \item[] Justification: The paper does not involve crowdsourcing nor research with human subjects.

\item {\bf Declaration of LLM usage}
    \item[] Question: Does the paper describe the usage of LLMs if it is an important, original, or non-standard component of the core methods in this research? Note that if the LLM is used only for writing, editing, or formatting purposes and does \emph{not} impact the core methodology, scientific rigor, or originality of the research, declaration is not required.
    \item[] Answer: \answerNA{} 
    \item[] Justification: The core method development in this research does not involve LLMs as any important, original, or non-standard components.

\end{enumerate}

\end{document}